\documentclass[letterpaper, 10 pt, conference]{ieeeconf}  

\IEEEoverridecommandlockouts                              

\usepackage[utf8]{inputenc} 
\usepackage[T1]{fontenc}    
\usepackage{hyperref}       
\usepackage{url}            
\usepackage{booktabs}       
\usepackage{amsfonts}       
\usepackage{nicefrac}       
\usepackage{microtype}      
\usepackage{xcolor}         

\usepackage{array}
\usepackage{graphicx}
\definecolor{lightblue}{rgb}{0.8,0.9,1}
\definecolor{darkblue}{rgb}{0.6, 0.75, 0.95}
\definecolor{aquamarine}{rgb}{0.5, 1.0, 0.83}
\definecolor{darklavender}{rgb}{0.8, 0.8, 0.98}
\usepackage{pifont}

\usepackage[dvipsnames]{xcolor}
\definecolor{darkgreen}{rgb}{0,0.5,0}
\newcommand{\greencheck}{\textcolor{darkgreen}{\ding{51}}}
\newcommand{\redx}{\textcolor{red}{\ding{55}}}

\usepackage{colortbl}
\usepackage{caption}
\usepackage{algorithm}
\usepackage{algorithmic}
\usepackage{amsmath}
\usepackage{soul}
\usepackage{subcaption}
\usepackage{booktabs}
\usepackage{multirow}
\usepackage{calc}
\usepackage{cite}

\DeclareMathOperator*{\argmax}{arg\,max}

\AtBeginEnvironment{quote}{\raggedright}

\usepackage{authblk}

\title{Personalized Embodied Navigation for Portable Object Finding}

\author[$1$]{Vishnu Sashank Dorbala*}
\author[$1$]{Bhrij Patel*}
\author[$2$]{Amrit Singh Bedi}
\author[$1$]{Dinesh Manocha\vspace{-0.2cm}}
\affil[$1$]{University of Maryland, College Park}
\affil[$2$]{University of Central Florida}

\newcolumntype{M}[1]{>{\centering\arraybackslash}m{#1}}

\newcommand\Item[1][]{%
  \ifx\relax#1\relax  \item \else \item[#1] \fi
  \abovedisplayskip=0pt\abovedisplayshortskip=0pt~\vspace*{-\baselineskip}}

\begin{document}

\maketitle
\thispagestyle{empty}
\pagestyle{empty}


\begin{abstract}
     Embodied navigation methods commonly operate in \textit{static} environments with \textit{stationary} objects. In this work, we present approaches for tackling navigation in dynamic scenarios with \textit{non-stationary} targets. In an indoor environment, we assume that these objects are everyday portable items moved by human intervention. We therefore formalize the problem as a personalized habit learning problem. To learn these habits, we introduce two \textbf{Transit-Aware Planning} (TAP) approaches that enrich embodied navigation policies with object path information. TAP improves performance in portable object finding by rewarding agents that learn to \textit{synchronize} their routes with target \textit{routes}. TAPs are evaluated on \textbf{Dynamic Object Maps (DOMs)}, a dynamic variant of node-attributed topological graphs with structured object transitions. DOMs mimic human habits to simulate realistic object routes on a graph. We test TAP agents both in simulation as well as the real-world. In the MP3D simulator, TAP improves the success of a vanilla agent by $\textbf{21.1\%}$ in finding non-stationary targets, while also generalizing better from static environments by $\textbf{44.5\%}$ when measured by Relative Change in Success. In the real-world, we note a similar $\textbf{18.3\%}$ increase on average, in multiple transit scenarios. We present qualitative inferences of TAP-agents deployed in the real world, showing them to be especially better at providing personalized assistance by finding targets in positions that they are usually not expected to be in (a toothbrush in a workspace). We also provide details of our \textit{real-to-sim} pipeline, which allows researchers to generate simulations of their own physical environments for TAP, aiming to foster research in this area.
     \footnote{ Code and data for our benchmark will be made publicly available upon acceptance.}
\end{abstract}

\section{Introduction}
\label{sec:intro}

Embodied navigation involves an agent moving in unseen physical or virtual environments to carry out human-guided tasks such as locating objects \cite{objnaveval, rudra2023contextual, objg1, objg3}, following instructions \cite{vlnbert, vlnhigh1, r2r, qiREVERIERemoteEmbodied2020}, or reaching a point in space \cite{pointnav3, ddppo}. State-of-the-art methods in this field use Reinforcement Learning (RL) ~\cite{ovrl2, ddppo, imagenavtopo} and Large Language Model (LLM) based schemes~\cite{shahLMNavRoboticNavigation2022, objg1} to achieve superior task performance, and these policies also extend to multi-object finding \cite{goat, multion}. Usually, these agents operate with only local observations and not from additional sources like external cameras \cite{objectnavrevisited, objg3}. 

However, a key feature of current embodied navigation paradigms~\cite{multion, goat, lgx, chenTopologicalPlanningTransformers2021} is their reliance on static graphs, where objects within nodes remain stationary. This static assumption is limiting, as real-world environments are inherently dynamic, where users frequently move portable objects from place to place over long periods, such as shifting their phone or wallet between rooms throughout the day. Studies on robots in collaborative environments ~\cite{habitstudy4, habitstudy2, habitstudy3} suggest that human object placements follow structured patterns shaped by routines and habits, introducing a degree of entropy in object locations over time. Despite the importance of non-stationary environments, existing navigation methods fail to model temporal variability, limiting their deployment in real-world settings.


\begin{figure}[t!]
    \centering
    \includegraphics[width=\linewidth]{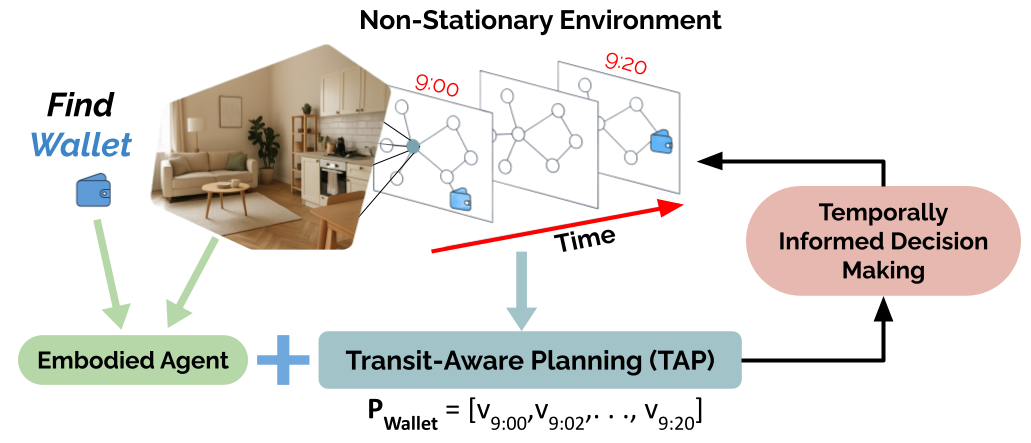}
    \caption{\textbf{Transit-Aware Planning (TAP)}: We introduce TAP to tackle embodied navigation in non-stationary environments. In this figure, an embodied agent is tasked with finding a wallet that changes positions between 9:00 and 9:20. $P_{\text{wallet}}$ represents the transit route of the wallet. TAP makes RL and LLM-based navigation agents ``\textit{transit-aware}'' by augmenting their policies for temporally informed decision-making.}
    \label{fig:TAP}
    \vspace{-0.5cm}
\end{figure}


In the ObjectNav task \cite{lgx, objnaveval}, for instance, SotA navigation approaches all involve finding \textit{stationary} targets using RL \cite{ovrl2, ddppo, objnaveval, sutton1998reinforcement} and LLM-based zero-shot methods \cite{lgx, shahLMNavRoboticNavigation2022, objgoalnav1}. RL policies for multi-object settings \cite{multion, goat} have also been proposed, but these are also based on static environments. In contrast, navigating to non-stationary targets that move as the agent moves necessitates the temporal modelling of object transit to guide the agent.

For example, OVRL-V2 \cite{ovrl2} uses CLIP text embeddings and ViT-processed RGB images, while DD-PPO \cite{ddppo} similarly relies on GPS readings and RGB-D sensors; both use only static scene observations. Their reward functions similarly assume target stationarity, with OVRL-V2 and DD-PPO penalizing absolute distance changes to the goal.
Using these approaches directly in non-stationary environments shows poor adaptability, and this can be attributed to a \textit{high-variance, noisy} signal hindering policy convergence.

For real-world deployment, these approaches must \textit{generalize}, i.e., maintain performance in dynamic scenes with shifting targets.
Prior embodied work describes generalizability in terms of measuring an agent's capacity to adapt to novel settings, but this usually refers to improved performance in \textit{unseen} environments or unique synthesized guidance language \cite{general1, gensur1, general2, clipnav}. Little work defines generalizability in the context of adapting to dynamic settings with non-stationary targets.

\begin{figure}[t]
    \centering
    \includegraphics[width=0.8\linewidth]{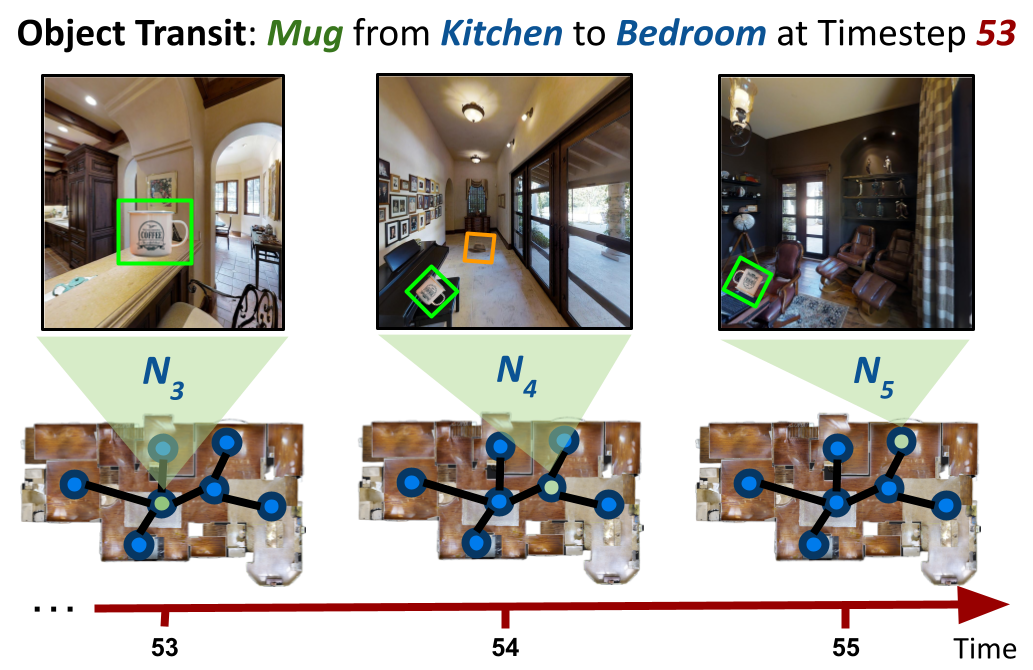}
    \caption{\textbf{Object Transition with Dynamic Object Maps}: Objects move at various timesteps in accordance with their natural rooms and transit scenarios. A mug is placed on a kitchen node $N_{3}$ at timestep $T=53$, but moves to a bedroom node $N_{5}$ at $T=55$ via node $N_{4}$. If an agent reaches the kitchen after $T=53$ (or the bedroom before $T=55$), it fails to see the mug. Multiple objects could also be at the same node (hat and mug in $N4$). 
    }
    \label{fig:portable_placement}
    \vspace{-0.5cm}
\end{figure}

\noindent \textbf{Main Results.} 
To address these issues, we present \textit{Transit-Aware Planning} (TAP) strategies that augment embodied navigation policies with temporal ``awareness'' on target positions (Fig. \ref{fig:TAP}). Rather than naively chasing target objects, TAP augmented agents learn to \textit{intercept} target routes for better convergence.
To evaluate TAP, we also introduce Dynamic Object Maps (DOMs), a dynamic variant of topological graphs incorporating \textit{structured} object transitions. Figure \ref{fig:portable_placement} illustrates this concept. DOM node attributes change over time, i.e. target objects move from node to node at each timestep, while the edges remain fixed. A successful agent learns to exploit structured object paths in the environment, adapting itself to better find non-stationary targets over time. Our work seeks to redefine success from traditional embodied navigation tasks, by emphasizing not just \textbf{where} to navigate but also \textbf{when} to get there. Our contributions are summarized as follows:

\noindent \textbf{Transit-Aware Planning (TAP):} We propose TAP, a set of novel strategies that equip embodied agents with \textit{non-stationary} target finding capabilities. TAP aims to adapt standard RL \cite{ddppo} and LLM \cite{lgx}- based navigation policies, initially developed for static target finding, to non-stationary targets via TAP-RL and TAP-LLM.

\noindent \textbf{Dynamic Object Maps (DOMs):} We propose a \textit{structured} formulation for object transitions on static topological graphs, transforming them into DOMs. Inspired by human habits, we define three \textbf{object transition scenarios} to govern the movement of portable objects in a scene, providing varying levels of entropy. DOMs apply to static TGs from datasets like Matterport3D (MP3D) \cite{matterport3d} and HM3D \cite{mp3dhabitat}. DOMs provide standardized environments to test high-level planning by learning object transition paths.
    
\noindent \textbf{Generalizability Benchmark}: We evaluate $6$ different navigation agents on a multi-object finding task across static and DOM environments. Our TAP-RL and TAP-LLM agents show an average improved Success Rate of $21.1\%$ and generalize better by $44.5\%$ in Relative Change in Success \cite{clipnav} over non-TAP counterparts. 

\noindent\textbf{Real-World Transfer:} Finally, we present a method to transfer our work into the real-world by augmenting scene graphs. We infer that TAP-enabled agents are much better at finding targets at unexpected or unique locations, enabling more personalized assistance over non-TAP counterparts.

\section{Related Works}



\noindent \textbf{Navigation in Dynamic Environments.} Traditionally, navigation in dynamic environments has focused on \textit{low-level} planning tasks, such as crowd avoidance \cite{densecavoid, dynamicenv1, dynamicenv2, park2013real, park2012itomp} and socially aware navigation \cite{social2, braids}, where agents continuously maneuver to avoid obstacles while minimizing trajectory length or time. In contrast, embodied navigation tasks employing \textit{high-level} planning on graphs are predominantly studied in static environments with stationary targets, with limited research on handling dynamic targets ~\cite{vlnhigh1, vlnhigh2}. In our work, we introduce dynamism on a graph-level, where agents must reason over \textit{long-term} spatio-temporal changes of objects for navigation decisions. While recent simulators like Habitat 3.0~\cite{habitat3} include object rearrangement tasks, these primarily focus on agents modifying the environment and not finding non-stationary targets on a graph. Planning in dynamic environments has been studied previously \cite{nonstationary1, nonstationary2, nonstationary3, petti2005safe, 933196}, with recent approaches even employing LLMs in conjunction with multi-arm bandits \cite{nonstationaryllm1}. These schemes usually propose memory augmentations with hierarchical planning procedures and frame the problem from an obstacle avoidance standpoint \cite{nonstationary1, nonstationary4, nonstationary5}. In contrast, our work considers when objects themselves are non-stationary. 

\noindent \textbf{Object Searching with Uncertainty.} Rudra et. al. \cite{rudra2023contextual} define small, portable objects around the house and propose a contextual bandit scheme that aims to learn the likelihood of finding an object at various waypoints. In their case, however, object locations are shuffled only after each episode, meaning it finally boils down to finding stationary targets at each timestep. In contrast, we tackle a \textit{truly} dynamic case, where objects move even \textit{during} the episode. This definition adds a layer of complexity as the embodied agent must now plan to navigate towards a shifting target, requiring an explicit modeling of the object's transit. Kurenkov et. al. \cite{kurenkov2023modeling} introduced dynamizing household environments and experimented with scene graph memory to predict object locations. They also performed target object-finding experiments, but the target was stationary as the agent moved. In our work, the objects move as the agent moves, and we deal with topological graphs, not scene graphs. Wang et. al \cite{wang2024dynamic} used LLM-generated human activities to dynamize a topological graph of a household environment and performed ObjectNav experiments in them. In contrast, our work presents transit-aware planning for portable object finding.





\section{Environment Formulation: Habitual Object Movement on Topological Graphs}



\noindent \textbf{Defining Transit Scenarios $\Lambda$:} Human habit formation via object placement can be attributed to \textit{cognitive offloading}, where individuals rely on the environment to reduce memory demands \cite{cogoff, habitform}. This reliance on the environment influences the spatial distribution of objects. Frequently used objects (like keys or phones) might move more flexibly across rooms (high entropy), while function-specific objects (like dumbbells or toothbrushes) might be more constrained (low entropy). We use this structured transition behaviour to define an object's \textit{transition entropy}, which we use to formulate $\Lambda$. We consider three object transit scenarios at decreasing levels of \textit{transit entropy}, \emph{---} \textbf{Random}, \textbf{Semi-Routine}, and \textbf{Fully-Routine}, 
outlined in Table \ref{tab:placement_cases}. 

\begin{table}[h!]
    \centering
    \small
    \begin{tabular}{lccc}
    \textbf{Scenario ($\mathbf{\Lambda}$)} & \textbf{Fixed Rooms} & \textbf{Fixed Paths} & \textbf{Entropy}\\
    \hline
    Static & \greencheck & N/A & None\\
    \hline
    Random & \redx & \redx & High\\ 
    \hline
    Semi-Routine & \greencheck & \redx & Medium\\
    \hline
    Fully-Routine & \greencheck & \greencheck & Low\\
    \hline
    \end{tabular}
    \vspace{0.4cm}
     \caption{\textbf{Object Transit Scenarios $\mathbf{\Lambda}$}: Portable targets transition in the graph under structured scenarios inspired by human habits. In the random case, the portable objects can move to any room at any time during each episode. In the routine cases, the rooms that the target objects can travel to are fixed. The static scenario is a baseline with stationary items i.e., zero entropy.}
    \label{tab:placement_cases}
    \vspace{-0.3cm}
\end{table}

\noindent \textbf{Dynamic Object Maps (DOMs):} To adapt static TG environments to test personalized, portable object finding, we define DOMs as node-attributed graphs where the attributes (objects) evolve. Let $G = (V, E)$ be a topological graph with nodes $V$ representing locations, and static edges $E$ are fixed paths between nodes. Let $\mathcal{O}$ be the set of portable objects on $G$. At each timestep $t$, $v \in V$ has a subset $O_{v}(t) \subseteq \mathcal{O}$ as attributes. Figure \ref{fig:portable_placement} illustrates a DOM where objects $\mathcal{O}$ move in $G$ based on $\Lambda$. Let $\mathcal{P}$ be the set of object paths that characterize the evolving node attributes. Then, $\mathcal{P}_{o} \in \mathcal{P}$ is a node sequence $[v_{1}, \dots v_{n}]$ denoting the position of object $o$ until time $n$. Given $G$, $\mathcal{O}$, and $\Lambda$, we generate a DOM to provide a setting to test methods for learning $\mathcal P$ needed for personalized assistance. 

\section{Method: Transit-Aware Planning (TAP)}

\label{sec:Tas}


Now that we have defined our environments with portable objects, we describe our methods to learn the paths $\mathcal{P}$ from partial observability, a scenario seen in embodied tasks like ObjectNav \cite{objectnavrevisited, rudra2023contextual}, for personalized decision-making. Many prior embodied navigation tasks work under the premise of ``\textit{static and stationary}'', where targets are assumed to stay at one location throughout the episode. This premise is also true with zero-shot LLM-based approaches \cite{shahLMNavRoboticNavigation2022, lgx}. With portable, non-stationary objects however, agents need to be made ``aware'' of target objects in transit to improve performance, and this involves integrating them with temporal knowledge of the scene. TAP agents to capture target transitions $\mathcal{P}$ in the environment as a learning objective. 

\noindent\textbf{TAP-RL Agent}:
We modify the observations, rewards, and actions on a standard RL-based embodied navigation policy.

\noindent\textit{Observations and Actions:}
To make the RL agent transit-aware, at every timestep $t \in T$, we provide a snapshot of all the portable object positions $\mathcal{P}[t]$. Note, we never provide the entire object paths to the agent, as our objective is for them to learn these transitions $\mathcal{P}$. We employ a \textit{maskable} action space to ensure the agent moves to a neighboring node of $v_c$. 

\noindent\textit{Reward Objective:} 
Standard RL-based embodied navigation agents are trained with a reward to penalize the agent for being too far from the target \cite{ddppo,ovrl2}: $\mathcal{R}_{\text{dist}}= - \alpha \parallel{v_{c} - v_{o}}\parallel$. This formulation works well with stationary targets, since an agent learns to steadily decrease its distance to $v_{o}$. In a non-stationary setting, however, $v_{o}$ keeps changing, and as a result, even though $R_{\text{dist}}$ is a dense reward, the reward signal fluctuates too much to enable meaningful learning. This fluctuation is reflected in the poor performance of DD-PPO \cite{ddppo} on DOMs and is discussed later in the results section.

For TAP-RL, we replace $\mathcal{R}_{\text{dist}}$ with a sparse \textit{interception} reward $\mathcal{R}_{I}$ that triggers when the agent intersects an object path $P_{o}$. The reward is the number of objects in $\mathcal{O}$ with that being the case. Formally, $R_I = \sum_{o \in \mathcal{O}} \mathbf 1_{v_c \in P_{o}}$, where $\mathbf 1$ is an indicator function.
This reward incentivizes the agent to \textit{synchronize} its path with that of the target, rather than greedily ``chasing'' a target to minimize distance.
$\mathcal{R}_{I}$ along with a standard target finding reward $\mathcal{R}_{f}$ allows for more informed agent decisions for finding portable objects.


\noindent\textbf{TAP-LLM Agent}: LLM-based navigation agents use \textit{commonsense reasoning} about observations to predict navigable actions. In LLM \cite{lgx} for instance, if the target is a \textit{`leaf-shaped bowl'}, the agent uses observations of regular objects like \textit{`table'} and \textit{`sofa'}, to determine a direction that is more likely to have the target (the \textit{`table'}).

Objects transit scenarios are grounded in common-sense (See Table \ref{fig:portable_placement} in the supplementary). passing previously seen regular objects to the LLM would help the agent make more transit-aware decisions. Since LLM agents are predominantly zero-shot and work directly during inference time with no training, we modify the system prompt to make them transit-aware. At each timestep $ts_{i}$, we gather the action sequence $a_{0->ts_{i}}$, and a set of objects in the scene (may include the target) $o_{t}$. In the subsequent timesteps, we pass this transit data $\mathcal{T}_{i} = [ts_{i}, a_{0->ts_{i}}, o_{t}]$ into the system prompt. A horizon is maintained to avoid token overflow by removing the earliest appended observation.


For both TAP-RL and TAP-LLM, $\Lambda$ plays an important role in determining agent success. An agent finding an object in random transit $\Lambda_{\text{random}}$ would find it a lot harder than one in cases $\Lambda_{\text{semi-routine}}$ or $\Lambda_{\text{routine}}$. For the former, even if an agent intersects during $\Lambda_{\text{random}}$ and follows the trajectory, it may never find the target as $\Lambda_{\text{random}}$ is spatio-temporally random.

\section{Task: Multi-Object Finding}
We focus on the task of finding multiple portable target objects over a fixed timespan.
This setting adds realism to standard multi-object finding tasks such as Multi-ON~\cite{multion} and GOAT~\cite{goat} where targets are stationary. In our dynamic setting, we argue that multi-object finding is preferable over finding a single portable object for two reasons: \textbf{1) Realistic Setting}: In real-world homes, users frequently move multiple objects simultaneously \cite{habitstudy2, habitstudy4}. Agents must track and infer object movements to provide assistance when asked, requiring monitoring multiple portable objects (e.g., phones, watches) rather than treating each search as an isolated event. \textbf{2) Lifelong Navigation}: A multi-object setup enables an agent to adapt its search policy in real-time, by using knowledge gathered from finding previous objects. An agent deployed at homes will likely have to face similar conditions, with a user asking to find multiple targets without `resetting' its position after each target, as with single-target tasks, similar to lifelong navigation\cite{wiyatno2022lifelong}.   

First, we formalize navigating on a DOM. Let $\Lambda$ describe the transitory motions of objects. Then, let $O$ be the set of portable objects found over $T$ timesteps for an agent starting at node $r$. Then $S = [r, T, \Lambda]$ represents a set of our experimental variables for this task. We then define our navigation objective as \textit{``Finding the \textbf{maximum} possible number of portable objects $O$ while traversing a dynamic environment represented by $S$''}. For downstream applications like embodied agents personalizing to households, finding multiple targets is a good task to study dynamic adaptability, due to dense reward signals \cite{goat}. The setting is also realistic as households have multiple people moving objects around simultaneously. We study the performance of different navigation policies with varying conditions $S$. 

In each episode, the agent tries to find as many portable objects as possible on a DOM in $T$ timesteps. Let $\tau$ be a finite trajectory of length $T$ representing a sequential list of visited nodes on the DOM, $G$, and let $\tau(t)$ be the $t$-th node in $\tau$. Now, define $O(\tau, t)$ as the set of portable objects found at $\tau(t)$. We define the set of portable objects found along $\tau$ as $O_\tau = \cup_{t=1}^T O(\tau, t)$. An agent with policy $\pi$ explores the DOM by generating a trajectory $\tau_{\pi}$, where each timestep $t$ corresponds to a selected node: $\tau_{\pi}(t) \in V$. We now can write the policy optimization problem as finding an optimal policy $\pi^*$ defined as $\pi^* = \argmax_{\pi}|O_{\tau_\pi}|$. Here, $\pi^*$ finds the most objects possible in $T$ timesteps. 



\captionsetup{font=small, labelfont=bf}

\begin{table*}[t]
\vspace{0.2cm}
\centering
\small
\renewcommand{\arraystretch}{1.1}
\setlength{\tabcolsep}{1.4pt}
\makebox[\textwidth][c]{
\begin{tabular}{lc cc >{\columncolor{lightblue}}c cc >{\columncolor{lightblue}}c cc >{\columncolor{lightblue}}c cc >{\columncolor{lightblue}}c}
\toprule
& & \multicolumn{3}{c}{$\Lambda$ \textbf{Random}} & \multicolumn{3}{c}{$\Lambda$ \textbf{Semi-Routine}} & \multicolumn{3}{c}{$\Lambda$ \textbf{Fully-Routine}} & \multicolumn{3}{c}{$\Lambda$ \textbf{Average}}\\ 
\cmidrule(lr){3-5}\cmidrule(lr){6-8}\cmidrule(lr){9-11}\cmidrule(lr){12-14}
\textbf{Policy} & \textbf{Metric (\%)}
& Static & DOM & \textbf{RCS} 
& Static & DOM & \textbf{RCS} 
& Static & DOM & \textbf{RCS}
& Static & DOM & \textbf{RCS} \\
\midrule
\multirow{2}{*}{Random} 
 & SR  & $42.0$ & $74.4$ &$+43.5$& $42.0$ &  $54.7$ & $+23.2$ & $42.0$ &  $48.3$ & $+13.0$ & $42.0$ & $59.1$ & $+29.0$ \\
 & TA & $33.5$ & $24.4$ & $-37.3$ & $33.5$ & $36.6$ & $+8.4$ & $33.5$ & $33.6$ & \cellcolor{darklavender} $\textbf{+0.29}$ & $33.5$ & $31.2$ & $-6.8$ \\
\midrule
\multirow{2}{*}{Oracle} 
 & SR  & $90.2$ & $78.4$ & $-13.0$ &  $90.2$ & $75.1$ & $-16.8$ &  $90.2$ & $55.9$ & $-38.0$ & $90.2$ & $69.8$ & $-22.6$\\
 & TA &  $45.6$ & $54.4$ & $+16.2$ & $45.6$ & $50.6$ & $+9.8$ & $45.6$ & $32.7$ & $-28.3$ & $45.6$ & $45.9$ & \cellcolor{darklavender}$\textbf{+0.6}$\\
\midrule
\multirow{2}{*}{RL \cite{ddppo}} 
 & SR  & $61.8$ & $33.4$ & $-45.9$ & $61.8$ & $21.8$ & $-64.7$ & $61.8$ & $12.2$ & $-80.1$ & $61.8$ & $22.4$ & $-63.7$ \\
 & TA & $59.8$ & $36.5$ & $-38.9$ & $59.8$ & $29.8$ & $-50.1$ & $59.8$ & $24.1$ & $-59.6$ & $59.8$ & $30.1$ & $-49.6$ \\
\midrule
\multirow{2}{*}{\textbf{TAP-RL}} 
 & SR  & $59.1$ & $55.2$ & $-6.6$ & $59.1$ & $57.8$ & \cellcolor{aquamarine} $\textbf{-2.3}$ & $59.1$ & $55.2$ & $-6.7$ & $59.1$ & $56.0$ & $-5.2$\\
 & TA & $41.6$ & $37.3$ & $-10.3$ & $41.6$ & $35.8$ & $-14.0$ & $41.6$ & $32.1$ & $-22.9$ & $41.6$ & $35.1$ & $-15.7$\\
\midrule
\multirow{2}{*}{LLM \cite{lgx}} 
 & SR & $28.5$ & $43.3$ & $+34.1$ & $28.5$ & $47.2$ & $+39.6$ & $28.5$ & $34.2$ & $+16.7$ & $28.5$ & $41.6$ & $+31.4$\\
 & TA & $24.3$ & $30.7$ & $+20.9$ & $24.3$ & $31.3$ & $+22.3$ & $24.3$ & $21.7$ & $-10.8$ & $24.3$ & $27.9$ & $+12.9$\\
\midrule
\multirow{2}{*}{\textbf{TAP-LLM}} 
 & SR & $50.6$ & $49.4$ & \cellcolor{aquamarine} $\textbf{-2.4}$ & $50.6$ & $53.2$ & $+4.8$ & $50.6$ & $47.9$ & \cellcolor{aquamarine} $\textbf{-5.4}$ & $50.6$ & $50.2$ & \cellcolor{aquamarine} $\textbf{-0.9}$\\
 & TA & $36.9$ & $33.8$ & \cellcolor{darklavender} $\textbf{-8.5}$ & $36.9$ & $34.6$ & \cellcolor{darklavender} $\textbf{-6.3}$ & $36.9$ & $32.4$ & $-12.2$ & $36.9$ & $33.6$ & $-9.0$\\
\bottomrule
\end{tabular}
}
\caption{\textbf{Navigation on DOMs}: We evaluate $6$ embodied navigation approaches on various DOMs to establish a benchmark. These values are used to compute RCS using the equation provided in \cite{clipnav}. An RCS value closer to $0\%$ indicates good generalizability to new conditions. Highlighted in \textcolor{darkgreen}{green} are the best RCS values for SR, while ones in \textcolor{violet}{violet} are the best values for TA. Observe the improved RCS scores of TAP-RL and TAP-LLM agents, indicating better generalizability to portable targets. Particularly note that the RL agent performs well in the static scenario, with a significant performance drop when subjected to DOMs. The LLM agent shows improved performance on DOMs, but we attribute this to multiple chance encounters with moving targets.}
\label{tab:nav_results}
\vspace{-0.3cm}
\end{table*}

\section{Experiments and Results}\label{sec:experiments}

We perform multi-object finding on the $4$ different $\Lambda$ scenarios defined in Table \ref{tab:placement_cases}. DD-PPO \cite{ddppo} and LLM \cite{lgx} are baseline RL and LLM agents and are enhanced as described in Section \ref{sec:Tas} to produce TAP-RL and TAP-LLM agents. We use GPT-4o for all LLM experiments. For PPO training, we use stable-baseline3\footnote{https://stable-baselines3.readthedocs.io/en/master/modules/ppo.html} package and its default parameters. We also benchmark with Random and Oracle agents. 

\subsection{DOM Setup}
We convert static topological graphs from the Matterport3D (MP3D) \cite{matterport3d} dataset to obtain DOMs. Each node contains panoramic images representing the scene, and edges represent the distance between them. The $\mathcal{O}$ and $\Lambda$ cases in our experiments are based on a pre-defined mapping between rooms and possible, portable objects. This mapping is provided in Table I in the supplementary, which is populated with commonsense knowledge on object locations. When generating the DOM, a pseudo-random seed $s$ helps differentiate the movement of portable objects per episode. Letting $e$ be the episode index, $s = e$ if $\Lambda$ is random or semi-routine. For the fully routine $\Lambda$, $s = 1$ across episodes, so the object transition routes remain fixed. 

We determine $4$ hyperparameters that influence performance i.e., $H = [r, E, T, \Lambda]$, where $r \in G$ is a random starting node chosen from $G$, $E$ is is the number of episodes per trial, $T$ is the number of timesteps per episode and $\Lambda$ is the transit scenario.
Thus, each agent is subjected to a total of $[Stationary (1) + Transit(3)] \times r \times E \times T$ experiments. We define each trial as a set of episodes over a fixed time from a starting node $r$. The maximum possible number of trials for a scan is thus equal to the size of $G$.
In our subset of $10$ MP3D scans, $|r|$ ranges from $[20, 215]$. For each scan, we set $|r| = 10$, $E = 20$, and $T = 30$. Although different objects can have different transition scenarios, we set all objects to have the same $\Lambda$ to better analyze the generalizability of a navigation scheme.

\noindent\textbf{Ground Truth Paths:}
Given a starting node $r \in G$, for each episode $e \in E$, there exist multiple optimal policies $\pi^* \in \Pi^*$ over $T$ generating ground truth paths $\tau_{\pi^*}$ that the agent could take for collecting the most objects. We compute these paths by brute force, simulating all possible trajectories from $r$ over $T$ timesteps, and use them in our evaluation.


\subsection{Evaluation Metrics}

\noindent \textbf{Success Rate (SR)}: For each episode, we measure the object-finding performance of an agent by dividing the number of objects found in that episode by the maximum possible number of objects that could have been found with an optimal policy $\pi^*$:  $SR= |O_{\tau_\pi}|/|O_{\tau_{\pi^*}}|$. This metric is similar to the Progress metric in \cite{multion}. \\
\noindent \textbf{Trajectory Alignment (TA)}:
To measure path efficiency, we modify the standard SPL metric \cite{anderson2018evaluation} to define TA as the maximum overlap between a ground-truth path and the agent's trajectory. Formally, $TA = \underset{\pi^* \in \Pi^{*}}{\max}|\tau_\pi \cap \tau_{\pi^{*}}|$.\\
\textbf{Relative Change in Success (RCS)}: We use RCS \cite{clipnav} to measure the generalizability from static to dynamic environments. It is the percent relative change between static and dynamic navigation performance. We measure RCS for various $\Lambda$ cases and report scores on SR and TA. Lower values indicate better generalizability, showing consistency in agent performance between static and dynamic environments.


\subsection{Navigation Inference}

\begin{figure}[h]
    \centering
    \includegraphics[width=\linewidth]{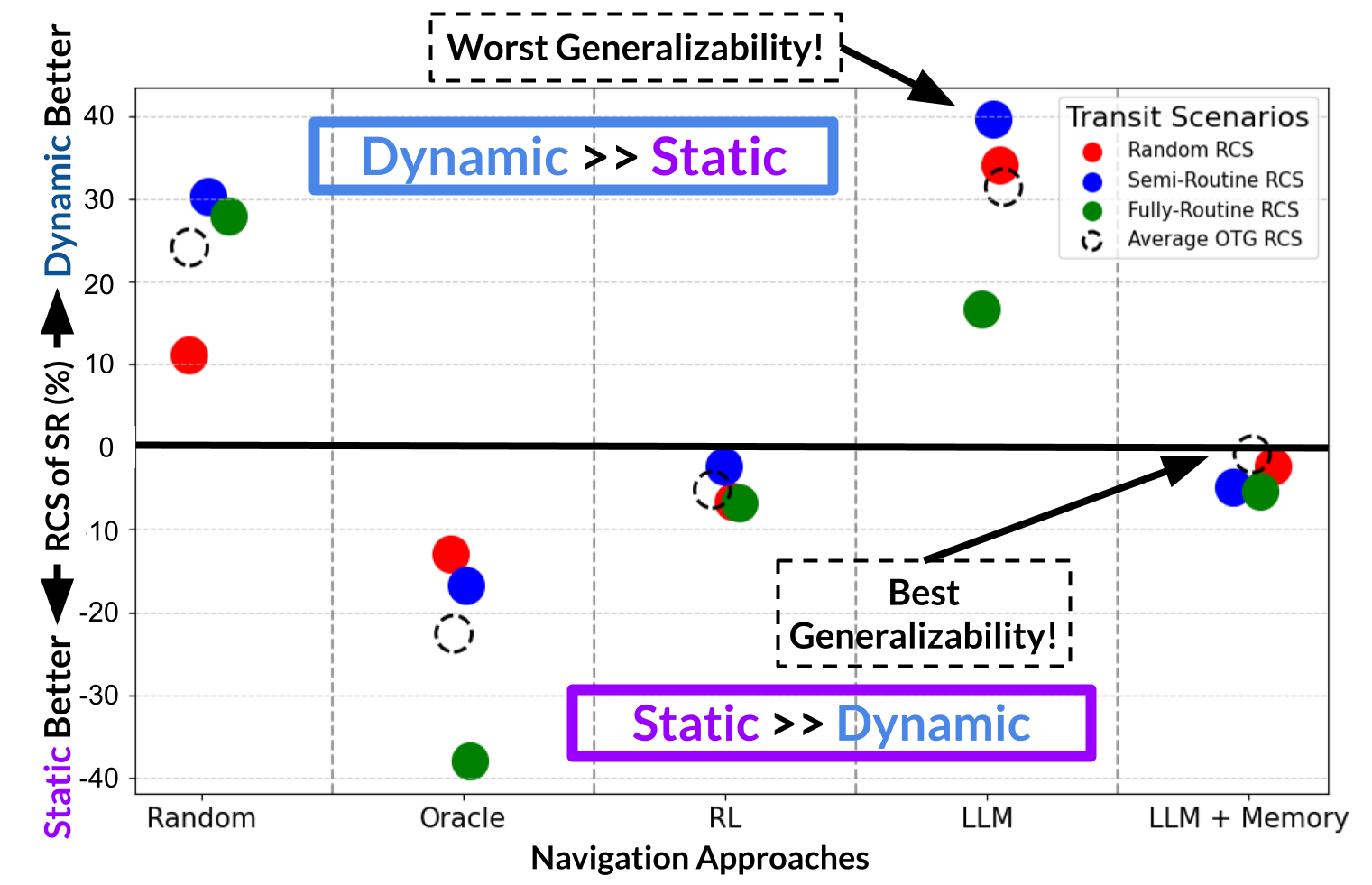}
    \caption{\textbf{Navigation Generalizability on DOMs:} Relative Change in Success (RCS) \cite{clipnav} measures an agent's adaptability, comparing dynamic to static TG performance (optimal RCS is $0\%$). Positive values indicate better dynamic performance, while negative values reflect poor adaptability. Both our TAP-enhanced agents approach optimal RCS, indicating better target finding in dynamic scenarios.}
    \label{fig:RCS_gen}
    \vspace{-0.3cm}
\end{figure}

\noindent\textbf{Main Results:} Table \ref{tab:nav_results} presents the results of our agents on the navigation task on DOMs. We draw $4$ key inferences to understand the generalizability of the navigation approaches:

\noindent\textbf{1) TAP improves Generalizability:}
Figure \ref{fig:RCS_gen} presents a comparison of all approaches on RCS. TAP-LLM and TAP-RL both have a relatively low RCS score, meaning that navigation agents perform comparably in stationary graphs and DOMs. This highlights the effectiveness of incorporating transit-awareness in portable objects. In contrast, the greedy heuristic Oracle agent and the zero-shot LLM lack transit information, with the former relying on privileged knowledge and latter utilizing commonsense cues about static observations for decision-making. We can infer TAP as vital for the real-world deployment of agents. Future work can explore the heuristic estimation of human habits as a prior for estimating transit paths for TAP.\\
\noindent\textbf{2) Zero-shot Methods Camp to Find Objects:} In the vanilla LLM approach, the agent finds more objects in the DOM than in the static environment, indicated by the higher DOM SR scores across all $\Lambda$s. From Table \ref{tab:nav_results}, in the Fully-Routine case for LLM, the RCS for SR is $+16.7\%$ but the RCS for TA is $-10.8\%$. These results suggest that while the LLM agent is finding more objects in DOMs than in static environments, its trajectories show less overlap with $\Pi^*$.
Upon inspection, we observe that this agent has learned to \textit{camp} between privileged nodes that are in many object routes.  
One possible explanation is that the LLM is relying solely on common-sense knowledge and goes to areas where most objects can be found, rather than trying to learn object routes. This challenge opens up research directions in DOM navigation reward design. 


\begin{figure*}[t!]
    \centering
    \includegraphics[width=0.85\linewidth]{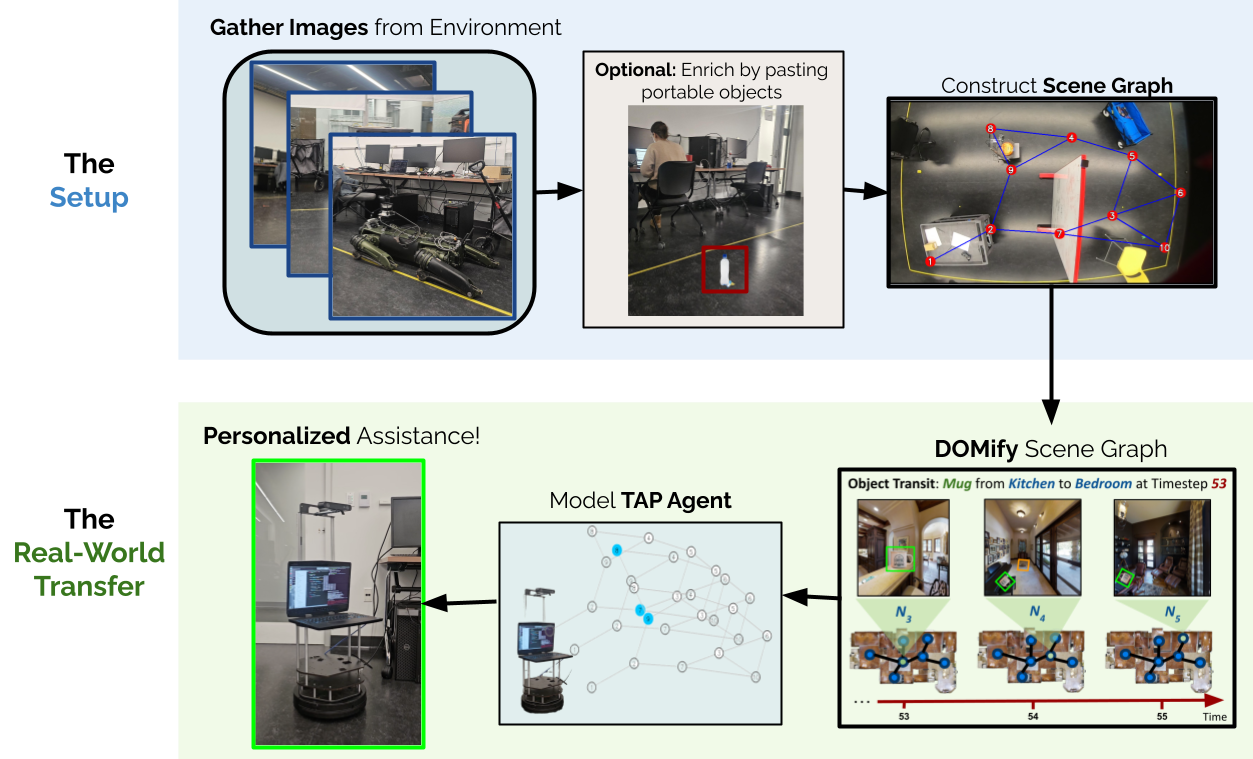}
    \caption{\textbf{Real World Transfer}: We present a real-to-sim pipeline for a real-world transfer of TAP agents. Since TAP agents learn habits over large periods of time, we create a simulation on a topological graph of the environment to enable us to gather statistics. The first phase involves gathering images from the scene and constructing a topological scene graph. We also optionally add small portable targets into these images to enrich this data (bottle in the top row, second image). We then DOMify these scene graphs, making it dynamic with moving objects, using Algorithm 1, and model a TAP agent with this data. Finally, we transfer this to the real-world for personalized assistance. Table \ref{tab:rw-stats} reports our results on \textbf{5} diverse environments. Please refer to the video attached that describes our pipeline and provides visuals.}
    \label{fig:real-world}
    \vspace{-0.4cm}
\end{figure*}

\noindent\textbf{3) Greedy Oracle Struggles on DOMs:} The Oracle agent uses a greedy heuristic that works well in $\Lambda_{\text{static}}$ by selecting a node towards the closest target object. On DOMs, however, it tends to take a suboptimal path because the agent keeps switching directions when another moving target object appears to be closer. This behavior is a direct approximation of the distance reward with DD-PPO that causes target chasing and highlights the need for transit awareness.

\noindent\textbf{4) Entropy Influences Performance}: In all policies barring the Random agent, the performance on a $\Lambda_{\text{routine}}$ exhibiting low entropy is usually worse than on a $\Lambda_{\text{semi-routine}}$ or $\Lambda_{\text{random}}$. This result can be attributed to the agent encountering more objects in environments with higher transit entropy. Despite this, we observe a strong performance of the TAP-RL agent in this case, with similar SR and TA values across all transit cases. This indicates that TAP can enhance embodied RL agents to become agnostic to scenarios where all target objects follow the same $\Lambda$. Future work could study a mixed $\Lambda$ with objects in the same DOM following distinct transit scenarios. 

\noindent \textbf{Ablation: Partial Observability with TAP-RL}: We ran an ablation experiment where we only provided the TAP-RL agent with the number of objects for the current adjacent nodes.
In terms of success rate (\%), we see degradation in performance in all DOM scenarios, $\Lambda_{\text{routine}}: 55.2 \rightarrow 12.6$, $\Lambda_{\text{semi}}: 57.8 \rightarrow 12.7$, and $\Lambda_{\text{random}}:  55.2 \rightarrow 14.3$, showing that an RL agent has better planning when given a larger observation space, indicating a need to use map-based planning with RL-based agents on DOMs.

\begin{figure*}[h!]
    \centering
    \begin{minipage}{0.48\linewidth}
        \centering
        \includegraphics[width=\linewidth]{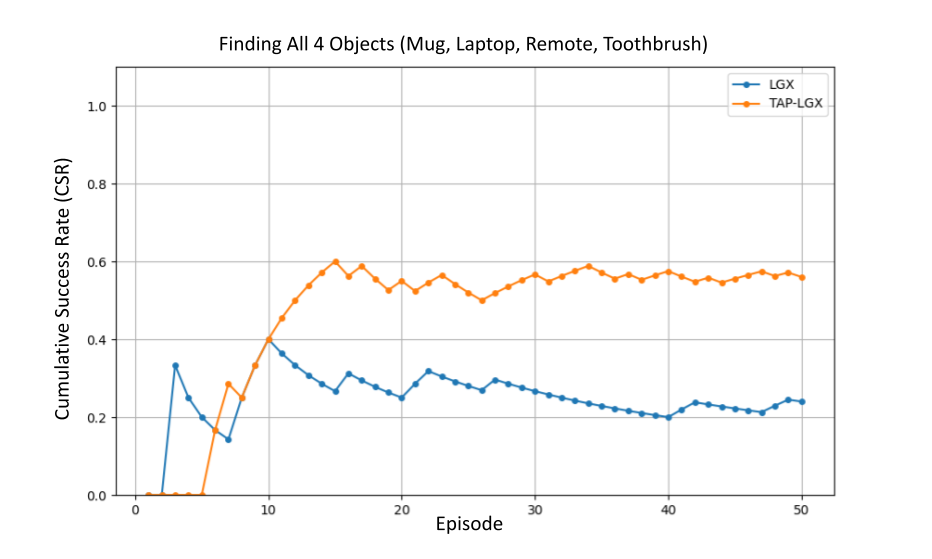}
        \caption{\textbf{Cumulative Success Rate (CSR)}: TAP-LLM outperforms the vanilla LLM overall in finding portable targets over time. Note the cumilitive success of finding these targets keeps increasing, since the TAP agent is made aware of transit behavior.}
        \label{fig:res_graph1}
    \end{minipage}
    \hfill
    \begin{minipage}{0.5\linewidth}
        \centering
        \includegraphics[width=\linewidth]{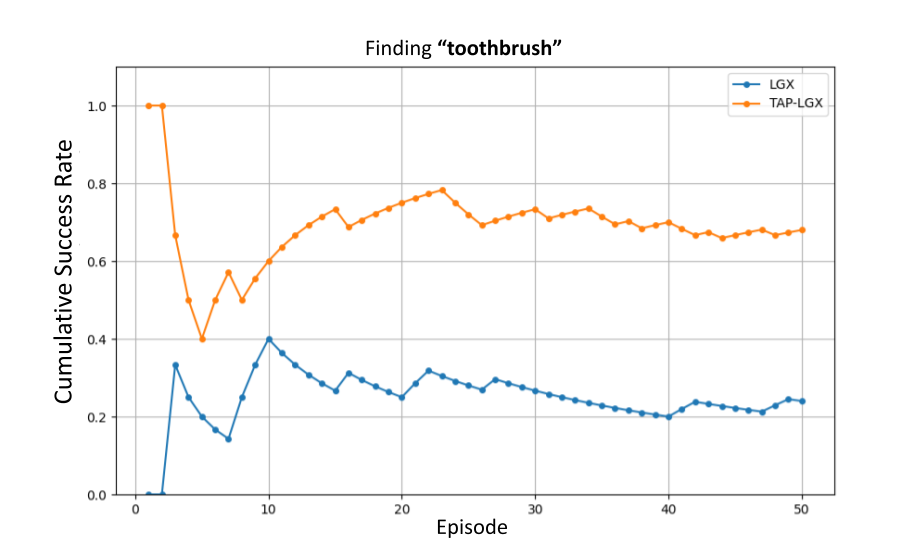}
        \caption{ \textbf{Non-Commonsense Target Success}: Observe the TAP-LLM agent can find non-commonsense targets like a toothbrush in our lab environment much better than its non-TAP counterpart. }
        \label{fig:res_graph2}
    \end{minipage}
\end{figure*}

\subsection{Discussion: Real-World Utility}

For a real-world agent to be accepted, it must slightly over-deliver on what it was intended to do \cite{brookslaws}. Learning user routines is an important capability in this regard, as it enables personalized decision-making in the absence of humans. 

\noindent\textbf{Real-World Lab Experiment}:
Our objective is to collect a scene graph of the real-world, make it dynamic by adding portable objects to it using DOMs, and run TAP agents. The entire process is highlighted in Figure \ref{fig:real-world}. We conduct real-world experiments as follows:

\begin{enumerate}
    \item \textbf{Scene Graph Construction}: For constructing a scene graph $\mathcal{S}$, we first collect images of our lab environment at different times of the day, and represent them across $10$ nodes. Each node represents a small location in our lab, containing multiple viewpoints of the scene. These viewpoints cover the periphery and middle spaces of our lab where people often move objects around. We run YOLO on these images and the detected objects of different shapes and sizes are the node attributes. Edges represent distances.
    \item \textbf{DOM Construction}: We place $4$ portable target objects across the nodes, as text descriptions for the LLM agent's system prompt: \textbf{white mug, laptop, black remote, toothbrush}. The white mug and laptop are \textit{commonsense} objects that are naturally present in labs. The \textit{black remote} and \textit{toothbrush} however, are out of place here, i.e., \textit{non-commonsense} targets. We purposefully choose these items to demonstrate the effectiveness of TAP agents for personalized target finding.
    Additionally, we gather more data samples that correspond to first-person views of the agent by enriching these images by pasting portable targets in them, and this is discussed in Figure \ref{fig:real-world}. Finally, we introduce shifting objects into the constructed scene graph to convert it into a DOM, as described in Algorithm 1. We use the routine case, where the target objects move in exactly the same way for each episode.
    \item \textbf{Run TAP Agent:}  We run both LLM and TAP-LLM to identify each of the $4$ targets, seeking to demonstrate the benefits of our TAP module in real-world settings. During each episode, the agent moves between nodes and samples a random viewpoint from the gathered data. For a vanilla non-TAP agent, we use LLM in its vanilla form, which purely operates on commonsense knowledge on objects detected to decide where to go. For the TAP agent, we additionally augment LLM with memory to keep track of timesteps, actions, and objects detected, and this info is passed on across episodes.
\end{enumerate}

\begin{figure}[t!]
    \centering
    \includegraphics[width=\linewidth]{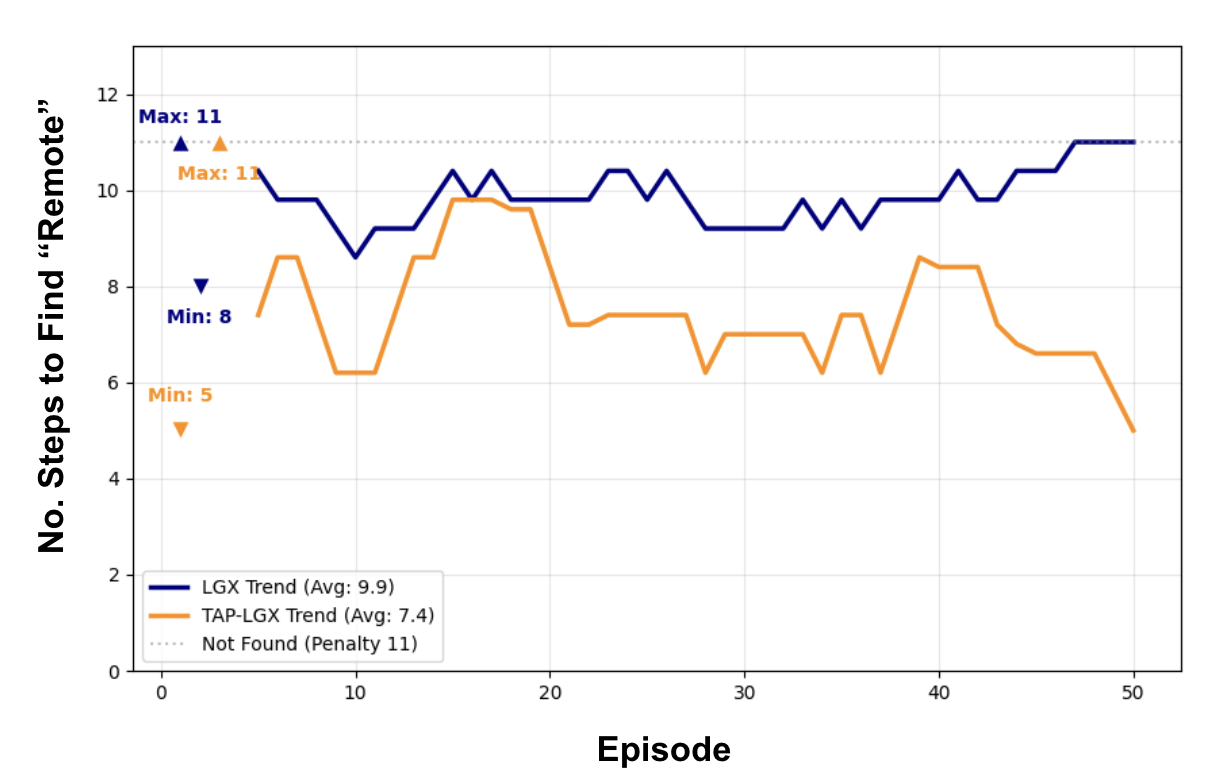}
    \caption{Number of steps to find the ``remote'' object in the lab environment shown in Figure \ref{fig:real-world}. Note the improved performance of TAP-LLM over time, in being able to find the remote much faster than the LLM agent. This highlights the influence of memory in enabling better non-stationary target finding over time.}
    \label{fig:res_graph3}
    \vspace{-0.4cm}
\end{figure}

\noindent\textbf{Inferences:} We report Cumulative Success Rate (CSR) values on $50$ episodes for the TAP-LLM and a vanilla LLM agent in graphs \ref{fig:res_graph1} and \ref{fig:res_graph2}. In graph \ref{fig:res_graph1}, we highlight the significant improvement of the TAP agent over time, in learning to find targets better than the vanilla LLM agent (LLM). Figure \ref{fig:res_graph2} presents the success of specifically the \textbf{toothbrush}, an unusually placed object, that is usually not present in office environments. Observe that TAP-LLM shows superior performance in finding this target that does not follow commonsense placement, since it can learn routine object paths over relying on just commonsense. Please refer to the new supplementary video attached for a detailed explanation and more visuals. We also plot in Figure \ref{fig:res_graph3} the number of steps to find the first target object, or \textbf{Time-To-Find (TTF)}. Figure \ref{fig:res_graph3} shows the TTF between LLM and TAP-LLM. Observe that TAP-LLM finds targets much faster over time. This results supports what Figure \ref{fig:res_graph1} shows, with TAP-LLM improving its success over time to find objects over the vanilla LLM agent.

The ``\textit{real-to-sim}'' transfer allows us to simulate multiple runs of TAP-LLM agents on this graph and validate the deployability of our agent in the real world. Further, it allows us to derive various statistics as we did with MP3D earlier. We run $50$ episodes to obtain success rates in various object transit scenarios. These are shown in Table \ref{tab:rw-stats}.

\begin{table}[t!]
    \centering
    \begin{tabular}{c|c|c|c}
         & Random $\Lambda$ (\%)& Semi-Routine $\Lambda$ (\%)& Routine $\Lambda$ (\%)\\       
        LLM & $37.5$ & $45.0$ & $57.5$ \\
        TAP-LLM & $\textbf{50.0}$ & $\textbf{65.0}$ & $\textbf{80.0}$
    \end{tabular}
    \caption{\textbf{Real World SR}: We report average success rates for finding portable objects in our lab environment over $50$ episodes. Note the improved performance of the TAP-LLM approach over a vanilla LLM scheme. Transit awareness helps improve performance in the routine cases, where paths have structure.}
    \label{tab:rw-stats}
    \vspace{-0.5cm}
\end{table}


\section{Conclusion, Limitations, and Further Work}
We present a novel approach to perform embodied navigation in dynamic environments with portable objects. Our method enhances RL and LLM-based policies traditionally modelled for stationary targets with a TAP strategy that enables them to capture object transit information. To evaluate our TAP-enhanced agents, we introduce DOMs that translate static topological graphs into dynamic ones with shifting objects. Object transitions in DOMs aim to simulate human object placement habits. Our work challenges the underlying \textit{``static graph, stationary target''} assumption common with embodied navigation literature, and presents a navigation benchmark in a dynamic environment with non-stationary, portable objects. On MP3D, our TAP-enhanced agents outperform non-TAP counterparts by $21.1\%$ when measured by average Success Rate in DOMs, while also drastically generalizing better to non-stationary environments by over $44.5\%$ when measured by RCS. We further conduct a real-world transfer of our approach, and note the superior performance of TAP agents on objects that are not placed in commonsense locations (eg. toothbrushes in labs). Future work could study mixed object transit cases and incorporate agents with higher degrees of freedom. Other directions could include incorporating more sophisticated measures of semantic memory to reduce the context length while maintaining dense information on human habits. While foundational knowledge can be help, it also can hurt generalizability across cultural semantics, e.g., some households call a ``\textit{flashlight}'' is called a ``\textit{torch}''.  Code and data will be made publicly available to foster research.

{\small
\bibliographystyle{IEEEtran}
\bibliography{refs}

}

\clearpage
\setcounter{page}{1}
\section*{Supplementary Material}

\begin{table*}[h]
\centering
\small 
\setlength{\tabcolsep}{10pt}
\begin{tabular}{l|p{0.83\textwidth}}
\hline
\textbf{Room} & \textbf{Portable Objects} \\
\hline
Bedroom & Charger, Water Bottle, Smartwatch, Laptop, Notebook, Toothbrush, Mug, Flash Drive, Phone, Headphones, Hat \\
\hline
Garage  & Screwdriver, Flashlight, Mug, Phone, Headphones, Hat \\
\hline
Dining & Salt and Pepper Shakers, Portable Speaker, Charger, Water Bottle, Mug, Bowl, Phone, Headphones, Hat \\
\hline
Office & Charger, Laptop, Hat, Notebook, Flash Drive, Mug, Phone, Headphones \\
\hline
Bathroom & Toothbrush, Phone, First-Aid Kit \\
\hline
Kitchen & Salt and Pepper Shakers, Hat, Mug, Bowl, Phone, Headphones, First-Aid Kit \\
\hline
Lounge & Playing Cards, Mug, Portable Speaker, Charger, Water Bottle, Laptop, Phone, Flash Drive, Dice, Headphones, Hat \\
\hline
Gym & Dumbbells, Jumprope, Smartwatch, Phone, Headphones, Hat \\
\hline
Outdoor & Jumprope, Smartwatch, Portable Speaker, Phone, Water Bottle, Headphones, Hat \\
\hline
Recreation & Playing Cards, Dice, Water Bottle, Headphones, Hat \\
\hline
\end{tabular}
\caption{\textbf{Rooms and Portable Objects Mapping for Experiments}: We map $21$ portable objects to a set of household rooms. We generate this mapping with GPT-3.5, asking it to provide us a list of rooms commonly found in a house and common objects you would typically find in them. This mapping is used to set destinations for object transit. During each episode, objects are placed in various rooms for a range of timesteps. Commonly moved objects such as \textit{phone, headphones} are associated with $9$ different rooms, while less commonly shifted ones such as \textit{dumbbells} appear only in the Gym.}
\label{tab:rooms_objects}
\vspace{-0.4cm}
\end{table*}

\section{DOMifying MP3D}
\label{app:dataset_detail}
We convert MP3D graphs into DOMs with the $21$ different objects presented in Table 3. On each of the $10$ scans chosen, we introduce routine, semi-routine and random object movement, and provide these as dictionaries with varying seeds. We will release these modified environments along with supporting code.

\begin{table}[h]
    \centering
    \small
    \setlength{\tabcolsep}{10pt}
    \caption{Summary statistics of scans used for Table \ref{tab:nav_results}, including each scan's ID, number of nodes, edges, and rooms. If a scan has multiple rooms of the same type (e.g. two bathrooms), each instance is counted separately from the total.}
    \vspace{0.2cm}
    \resizebox{1\columnwidth}{!}{\begin{tabular}{|c|c|c|c|}
    \hline
        \textbf{Scan ID} & \textbf{\# of Nodes} & \textbf{\# of Edges} & \textbf{\# of Rooms} \\
        \hline
        QUCTc6BB5sX & 145 & 248 & 28 \\
        \hline
        8194nk5LbLH & 20 & 32 & 6\\
        \hline
        TbHJrupSAjP & 114 & 221 & 28\\
        \hline
        2azQ1b91cZZ & 215 & 531 & 30 \\
        \hline
        oLBMNvg9in8 & 111 & 185 & 31 \\
        \hline
        zsNo4HB9uLZ & 53 & 84 & 17 \\
        \hline
        EU6Fwq7SyZv & 78 & 166 & 19 \\
        \hline
        X7HyMhZNoso & 84 & 143 & 25 \\
        \hline
        x8F5xyUWy9e & 43 & 86 & 10 \\
        \hline
        Z6MFQCViBuw & 58 & 91 & 18 \\
        \hline
    \end{tabular}}
    \label{tab:scan_details}
\end{table}

\subsection{Implementing Object Placement}

In this section, we give more details on how we convert Matterport3D environments into DOMs.

\noindent\textbf{Matterport3D Modifications:}

Each Matterport3D (MP3D) scan represents a household environment consisting of a set of panoramic view points. Along with the viewpoints (or nodes), we are also given the exact 3D position as well as the relative distance between them.
For modifying the MP3D environment, we first construct topological graphs of each scan, with nodes containing the position and the panoramic image, and edges containing relative distance between them. We consider 10 different scans chosen from the REVERIE \cite{qiREVERIERemoteEmbodied2020} and R2R \cite{r2r} unseen validation splits for inference. We choose scans according to these datasets as they contain a variety of rooms for us to populate (Refer Table \ref{tab:rooms_objects} in the main paper).

The nodes of the topological graphs are then updated at each timestep with the portable objects according to the object placement scenario that has been chosen.

\noindent\textbf{Strategy Overview:}

We compute trajectories of the portable objects offline. For a fixed random seed $s$ and for each portable object $o_p \in O_p$, we create a sequence of nodes from the graph. In the random transit scenario, we choose any node in the graph. In the routine and semi-routine scenarios, we only choose nodes from plausible rooms. If the node chosen is not the same as the current node, we find the shortest path between the current node and the target node and add the nodes in the path to the sequence representing the trajectory. Once $o_p$ gets to the target node, it stays there for either $2$ or $3$ timesteps, determined by a random number generator. For each timestep $o_p$ stays at the node, we add the node to the trajectory sequence. After the staying period is over we select a new node and repeat the process until we hit $T$ timesteps. 

We take the resulting trajectories and restructure the data to model an evolving graph over a period of $T$ timesteps. The resulting structure is a nested dictionary representing the changing graph. The keys of the outer dictionary are the timesteps [$1, ..., T$]. The inner dictionary for each timestep $t$ has the node string ids as the keys and the values are the list of portable objects at that node at $T$. At each timestep $t \in [1, ..., T]$ when the agent reaches a node, we simply use $t$ and the node id to retrieve the list of portable objects the agent is currently observing.





\subsection{TAP-PPO}
We modify a DD-PPO agent with temporal information about target objects in its observation space. As mentioned in section \ref{sec:Tas}, we gather object paths $\mathcal{P}$ for augmenting this space. 

In our task, since we treat each object equally, i.e. the objective is simply to maximize finding targets, we include $\mathcal P$ as a single list $l = [v_{1}, v_{2},\dots,v_{n}]$ of size $n = |G|$. Each index of $l$ corresponds to a different node in $G$. For each index $i$ of $l$, $v_i$ is the number of unique target objects at the $i$-th node in $G$. $l$ is updated at each timestep, and passed to the observation space along with the agent's current position and timestep. The agent's action space is masked, with the entire graph being the action space, and the mask allowing only for selecting neighboring nodes.

Additionally, we swap out the distance reward $R_{dist}$ with the intersection reward $R_{I}$ described in the main text. $R_{I}$ here is triggered whenever the agent encounters a node where a target object has been in the last $5$ timesteps. We do not make any other modifications to the default setup.

\subsection{TAP-LLM}

We utilize the official implementation of LGX \footnote{\url{https://github.com/vdorbala/LGX}} and incorporate it into our modified MP3D environment. Briefly, at each timestep, LGX scans the node for objects, and asks an LLM for directions to reach a target. In our case, we seek to maximize finding portable targets. As such, we prompt GPT-4 with the following base prompts - 

\begin{quote}
 \textbf{System Prompt} - ``I am a smart robot trying to find as many portable objects as I can at home.'' \\
 \textbf{User Prompt} - ``Which object from \textit{$<$OBJECT\_LIST$>$} should I go towards to find a new portable object? Reply in ONE word.''
\end{quote}

The $<$OBJECT\_LIST$>$ here contains a set of objects that have been detected using YOLO-v8, as proposed in LGX. Additionally, we if a portable object is present at a certain node at a given timestep, we add it to this list.

The LLM then predicts a target object from the list, which is mapped to an adjacent node using our customized MP3D functions. The agent then moves to that node.

For the TAP-LGX case, we modify the System Prompt, with memory as presented in Section \ref{sec:Tas}. We additionally ask -
\begin{quote}
    \textbf{System Prompt}:\\
    ``I have seen the following objects and taken the following actions so far - \\
    1. $<$OBJECT\_LIST$>$: ACTION \\
    2. $<$OBJECT\_LIST$>$: ACTION \\
    \dots'' \\
    \textbf{User Prompt}:\\ 
    ``Which object from \textit{$<$OBJECT\_LIST$>$} should I go towards to find a new portable object? Reply in ONE word.''
\end{quote}

Note that we maintain a horizon to avoid token overflow, where the earliest appended observation is removed when the context length limit is reached.

\section{Video, Code and Dataset}

We provide an anonymous link to our Code: \url{https://anonymous.4open.science/r/otg-1AC0}.
The dataset will be present as a downloadable link in the code repository.

Furthermore, please see the attached video in the zip file for more information and demonstrations.


\end{document}